# ENERGETICS OF THE BRAIN AND AI

Anders Sandberg

Sapience Project 2016

## Synopsis

Does the energy requirements for the human brain give energy constraints that give reason to doubt the feasibility of artificial intelligence? This report will review some relevant estimates of brain bioenergetics and analyze some of the methods of estimating brain emulation energy requirements. Turning to AI, there are reasons to believe the energy requirements for *de novo* AI to have little correlation with brain (emulation) energy requirements since cost could depend merely of the cost of processing higher-level representations rather than billions of neural firings. Unless one thinks the human way of thinking is the most optimal or most easily implementable way of achieving software intelligence, we should expect *de novo* AI to make use of different, potentially very compressed and fast, processes.

ACM Computing Classification System (CCS): Hardware~Cellular neural networks • Hardware~Emerging technologies~Biology-related information processing~Neural systems

London, United Kingdom

Sapience Project is a thinktank dedicated to the study of disruptive and intelligent computing. Its charter is to identify, extrapolate, and anticipate disruptive, long-lasting and possibly unintended consequences of progressively intelligent computation on economy and society; and to syndicate focus reports and mitigation strategies.

## Board

Vic Callaghan, University of Essex

B. Jack Copeland, University of Canterbury

Amnon H. Eden, Sapience Project

Jim Moor, Dartmouth College

David Pearce, BLTC Research

Steve Phelps, Kings College London

Anders Sandberg, Oxford University

Tony Willson, Helmsman Services

Recently there has been both major enthusiasm for artificial intelligence (AI) and concerns that it might pose major risks to humanity (Bostrom 2014). While a number of high-profile researchers think AI safety should be a high priority (Future of Life Institute 2015), there is also significant disagreement about how much risk AI poses. This is especially true for questions about human-level and beyond AI.

Lawrence Krauss (Krauss 2015) is not worried about AI risk; while much of his complacency is based on a particular view of the trustworthiness and level of common sense exhibited by possible future AI that is pretty impossible to criticise, he makes a particular claim:

> *First, let's make one thing clear. Even with the exponential growth in computer storage and processing power over the past 40 years, thinking computers will require a digital architecture that bears little resemblance to current computers, nor are they likely to become competitive with consciousness in the near term. A simple physics thought experiment supports this claim:*
>
> *Given current power consumption by electronic computers, a computer with the storage and processing capability of the human mind would require in excess of 10 Terawatts of power, within a factor of two of the current power consumption of all of humanity. However, the human brain uses about 10 watts of power. This means a mismatch of a factor of $10^{12}$, or a million million. Over the past decade the doubling time for Megaflops/watt has been about 3 years. Even assuming Moore's Law continues unabated, this means it will take about 40 doubling times, or about 120 years, to reach a comparable power dissipation. Moreover, each doubling in efficiency requires a relatively radical change in technology, and it is extremely unlikely that 40 such doublings could be achieved without essentially changing the way computers compute.*

This claim has several problems. First, few, if any, AI developers think that we must stay with current architectures. Second, more importantly, the community concerned with superintelligence risk is generally agnostic about how soon smart AI could be developed: it does not have to happen soon for us to have a tough problem in need of a solution, given how hard the AI value alignment problem seems to be. Third, consciousness is likely irrelevant for instrumental intelligence; maybe the word is just used as a stand-in for some equally messy term like "mind", "common sense" or "human intelligence".

The interesting issue is however, what energy requirements and computational power tells us about human and machine intelligence, and vice versa. If energy is a major constraint on cognition then we have a way of constraining predictions and claims about future artificial minds.

# Computer and brain emulation energy use

I have earlier looked at the energy requirements of the Singularity (Sandberg 2015). To sum up, current computers are energy hogs requiring 2.5 TW of power globally, with an average cost around 25 nJ per operation. More efficient processors are certainly possible (many of the ones in current use are old and suboptimal). For example, current GPUs consume about a hundred Watts and have $10^{10}$ transistors, reaching performance in the 100 Gflops range, one nJ per flop or better[1]. Koomey's law states that the energy cost per operation halves every 1.57 years (not 3 years as Krauss says (Koomey et al. 2011)). So far, the growth of computing capacity has grown at about the same pace as energy efficiency, making the two trends cancel each other. In the end, Landauer's principle gives a lower bound of $kT \ln 2$ J per irreversible operation (R. Landauer 1961); one can circumvent this by using reversible or quantum computation, but there are costs to error correction — unless we use extremely slow and cold systems in the current era computation will be energy-intensive.

I am not sure what brain model Krauss bases his estimate on, but 10 TW/25 nJ = $4 \cdot 10^{20}$ operations per second (using slightly more efficient GPUs ups it to $10^{22}$ flops). Looking at the estimates of brain computational capacity in appendix A of my old roadmap (Sandberg and Bostrom 2008), this is higher than most. The only estimate that seem to be in the same ballpark is (Thagard 2002), which argues that the number of computational elements in the brain are far greater than the number of neurons (possibly even individual protein molecules). This is a strong claim, to say the least. Especially since current GPUs can do a somewhat credible job of end-to-end speech recognition and transcription (Catanzaro 2015): while that corresponds to a small part of a brain, it is hardly $10^{-11}$ of a brain.

Generally, assuming a certain number of operations per second in a brain and then calculating an energy cost will give you any answer you want. There are people who argue that what really matters is the tiny conscious bandwidth (maybe 40 bits/s or less) and that over a lifetime we may only learn a gigabit (T. K. Landauer 1986). I used $10^{22}$ to $10^{25}$ flops just to be on the safe side in one post (Sandberg 2009). AIimpacts.org has collected several estimates, getting the median estimate $10^{18}$ (Grace 2015b). They have also argued in favor of using TEPS (traversed edges per second) rather than flops, suggesting around $10^{14}$ TEPS for a human brain — a level that is soon within reach of some systems (Grace 2015a).

There are many apples-to-oranges comparisions here. A single processor operation may or may not correspond to a floating-point operation, let alone to what a GPU does or a TEPS. However, since we are in the land of order-of-magnitude estimates this may not matter too much.

# Brain energy use

We can turn things around: what does the energy use of human brains tell us about their computational capacity?

Ralph Merkle calculated back in 1989 that given 10 Watts of usable energy per human brain, and that the cost of each jump past a node of Ranvier costs $5 \cdot 10^{-15}$ J, producing $2 \cdot 10^{15}$ such operations. He estimated this was about equal to the number of synaptic operations, ending up with $10^{13}$–$10^{16}$ operations per second (Merkle 1989).

A calculation due to Karlheinz Meier argued the brain uses 20 W power, has 100 billion neurons firing per second, uses $10^{-10}$ J per action potential, plus it has $10^{15}$ synapses receiving signals at about 1 Hz, and uses $10^{-14}$ J per synaptic transmission. One can also do it from the bottom to the top: there are $10^9$ ATP molecules per action potential, $10^5$ are needed for synaptic transmission. $10^{-19}$ J per ATP gives $10^{-10}$ J per action potential and $10^{-14}$ J per synaptic transmission. Both these estimates converge on the same rough numbers, used by Meier to argue that we need much better hardware scaling if we ever want to get to this level of detail.

---

[1] The Nvidia Titan X reaches 6.2 Tflops using 250 W, 0.04 nJ per flop.

Digging deeper into neural energetics, maintaining resting potentials in neurons and glia account for 28% and 10% of the total brain metabolic cost, respectively, while the actual spiking activity is about 13% and transmitter release/recycling plus calcium movement contributes about 1% (Lennie 2003). Note how this is not too far from the equipartition in Meier's estimate. Looking at total brain metabolism this constrains the neural firing rate: more than 3.1 spikes per second per neuron would consume more energy than the brain normally consumes (and this is likely an optimistic estimate). The brain simply cannot afford firing more than 1% of neurons at the same time, so it likely relies on rather sparse representations.

Unmyelinated axons require about 5 nJ/cm to transmit action potentials (Crotty, Sangrey, and Levy 2006). In general, the brain gets around it through some current optimization (Alle, Roth, and Geiger 2009), myelinisation (which also speeds up transmission at the price of increased error rate), and likely many clever coding strategies. Biology is clearly strongly energy constrained. In addition, cooling a 20 W brain through a bloodflow of 750-1000 ml/min is relatively tight given that the arterial blood is already at body temperature.

20 W divided by $1.3 \cdot 10^{-21}$ J (the Landauer limit at body temperature) suggests a limit of no more than $1.6 \cdot 10^{22}$ irreversible operations per second. While a huge number, it is just a few orders higher than many of the estimates we have been juggling so far. If we say these operations are distributed across 100 billion neurons (which is at least within an order of magnitude of the real number (Azevedo et al. 2009)) we get 160 billion operations per second per neuron; if we instead treat synapses (about 8000 per neuron) as the loci we get 20 million operations per second per synapse.

Running the full Hodgkin-Huxley neural model at 1 ms resolution requires about 1200 flops, or 1.2 million flops per second of simulation per compartment (Izhikevich 2004). If we treat a synapse as a compartment, that is just 16.6 times the Landauer limit: if the neural simulation had multiple digit precision and erased a few of them per operation we would bump into the Landauer limit straight away. Synapses are actually fairly computationally efficient! At least at body temperature: cryogenically cooled computers could of course do much better. And as Izikievich, the originator of the 1200 flops estimate, likes to point out, his model requires just 13 flops (Izhikevich 2004): maybe we do not need to model the ion currents like Hodgkin-Huxley to get the right behavior, and can hence shave off two orders of magnitude.

# Information dissipation in neural networks

Just how much information is lost in neural processing?

A brain is an autonomous dynamical system changing internal state in a complicated way (let us ignore sensory inputs for the current discussion). If we start in a state somewhere within some predefined volume of state-space, over time the state will move to other states and the initial uncertainty will grow. Eventually the possible volume we can find the state in will have doubled, and we will have lost one bit of information.

Things are a bit more complicated, since the dynamics can contract along some dimensions and diverge along others: this is described by the Lyapunov exponents. If the trajectory has exponent $\lambda$ in some direction nearby trajectories diverge like $|x_a(t) - x_b(t)| \propto |x_a(0) - x_b(0)|e^{\lambda t}$ in that direction. In a dissipative dynamical system, the sum of the exponents is negative: in total, trajectories move towards some attractor set. However, if at least one of the exponents is positive, then this can be a strange attractor that the trajectories endlessly approach, yet they locally diverge from each other and gradually mix. So if you can only measure with a fixed precision at some point in time, you can not tell with certainly where the trajectory was before (because of the contraction due to negative exponents has thrown away starting location information), nor exactly where it will be on the attractor in the future (because the positive exponents are amplifying your current uncertainty).

A measure of the information loss is the Kolmogorov-Sinai entropy (Sinai 2009), which is bounded by $K \leq \sum_{\lambda_i > 0} \lambda_i$, the sum of the positive Lyapunov exponents. Therefore, if we calculate the KS-entropy of a neural system, we can estimate how much information is being thrown away per unit of time (and hence get a hint of energy dissipation).

Monteforte and Wolf looked at one simple neural model, the theta-neuron (Monteforte and Wolf 2010). They found a KS-entropy of roughly 1 bit per neuron and spike over a large range of parameters. Given the above estimates of about one spike per second per neuron, this gives us an overall information loss of $10^{11}$ bits/s in the brain, which is $1.3 \cdot 10^{-10}$ W at the Landauer limit — by this account, we are some 11 orders of magnitude away from thermodynamic perfection. In this picture, we should regard each action potential corresponding to roughly one irreversible yes/no decision: perhaps a not too unreasonable claim.

It is worth noticing that one can look at cognition as a system with a large-scale dynamics that has one entropy (corresponding to shifting between different high-level mental states) and microscale dynamics with different entropy (corresponding to the neural information processing). It is a safe bet that the biggest entropy costs are on the microscale (fast, numerous simple states) than the macroscale (slow, few but complex states). If a computing system has a dynamics mimicking the brain on the microscale, we should expect it to dissipate more than $10^{-10}$ W. This is a *very* weak bound.

# Energy of AI

Where does this leave us in regards to the energy requirements of artificial intelligence?

Assuming the same amount of energy is needed for a human and machine to do a cognitive task is a mistake.

First, as the Izikievich neuron demonstrates (Izhikevich 2004), it might be that judicious abstraction easily saves two orders of magnitude of computation/energy. Special purpose hardware can also save one or two orders of magnitude (Krizhevsky, Sutskever, and Hinton 2012; Lu et al. 2015; Park et al. 2015). Using general-purpose processors for fixed computations is very inefficient (this is why GPUs are at present so useful for neural processing: in many cases, we just want to perform the same action on many pieces of data rather than different actions on the same piece). Further careful design may reduce the demands even more (Cao, Chen, and Khosla 2014; Panda, Sengupta, and Roy 2015).

More importantly, on what level the task is implemented matters. Sorting or summing a list of a thousand elements is a fast computer operation that can be done in memory in microseconds, but is an hour-long task for a human. Because of our mental architecture we need to represent the information in a far more redundant and slow way, not to mention perform individual actions on the seconds time-scale. A computer sort uses a tight representation more like our low-level neural circuitry. I have no doubt one could string together biological neurons to perform a sort or sum operation quickly (c.f. (Ditto 2003)), but cognition happens on a higher, more general level of the system (intriguing speculations about idiot savants aside).

While we have reason to admire brains, they are also unable to perform certain very useful computations. In artificial neural networks we often employ non-local matrix operations like inversion to calculate optimal weights (Toutounian and Ataei 2009): these computations are not possible to perform locally in a distributed manner. Gradient descent algorithms such as backpropagation are unrealistic in a biological sense, but clearly very successful in deep learning. There is no shortage of papers describing various clever approximations that would allow a more biologically realistic system to perform similar operations — in fact, the brains may well be doing it — but artificial systems can perform them *directly*, and by using low-level hardware intended for it, very efficiently.

When a deep learning system learns object recognition in a week[2] it beats a human baby by many months. When it learns to do analogies from 1.6 billion text snippets in 85 minutes (Pennington, Socher, and Manning 2014) it beats human children by years. These are small domains, yet they are domains that are very important to humans and presumably develop as quickly and efficiently as possible in children.

Biology has many advantages in robustness and versatility, not to mention energy efficiency. Nevertheless, it is also fundamentally limited by what can be built out of cells with a particular kind of metabolism, the fact that organisms need to build themselves from the inside, and the need of solving problems that exist in a particular biospheric environment.

---

[2] The time to train a deep neural network on the ImageNet database on individual computers with GPU support is on the order of days (Liu 2016).

# Conclusion

Unless one thinks the human way of thinking is the most optimal or most easily implementable way, we should expect *de novo* AI to make use of different, potentially very compressed and fast, processes[3]. Hence, the costs of brain computation is merely a proof of existence that there are systems that effective. The same mental tasks could well be done by far less or far more efficient systems.

In the end, we may try to estimate fundamental energy costs of cognition to bound AI energy use. If human-like cognition takes a certain number of bit erasures per second, we would get some bound using Landauer (ignoring reversible computing). However, as the above discussion has showed, it may be that the actual computational cost needed is just some of the higher-level representations rather than billions of neural firings: until we actually understand intelligence, we cannot say. By that point the question is moot anyway.

Many people have the intuition that the cautious approach is to always state "things won't work". However, this mixes up cautious with conservative (or even reactionary). A better cautious approach is to recognize that "things *may* work", and then start checking the possible consequences. If we want a reassuring constraint on why certain things cannot happen it need to be tighter than energy estimates.

## Acknowledgments

An earlier version of this essay was previously published on my personal blog. Thanks to Amnon Eden for adapting it to a whitepaper, to Luke Muelhauser (2015) and Carl Shulman for originally pointing out Lawrence Krauss essay, and to Lawerence Krauss for stimulating writing this essay. I would also like to thank Eric Drexer and Katja Grace for useful comments.

---

[3] Brain emulation makes sense if one either cannot figure out how else to achieve AI, or one wants to copy extant brains for their properties or individuality.